\documentclass[11pt]{article}
\usepackage{nodalida2019}
\usepackage{times}
\usepackage{url}
\usepackage{latexsym}
\usepackage[utf8]{inputenc}
\usepackage{multirow}

\usepackage{xcolor}
\usepackage{fancyvrb}

\aclfinalcopy 

\title{Template-free Data-to-Text Generation of Finnish Sports News}

\author{Jenna Kanerva{\normalfont,} Samuel R\"onnqvist{\normalfont,} Riina Kekki{\normalfont,} Tapio Salakoski {\normalfont and} Filip Ginter \\
  TurkuNLP \\
  Department of Future Technologies \\
  University of Turku, Finland \\
  {\tt \{jmnybl,saanro,rieeke,sala,figint\}@utu.fi} \\
  }

\date{}

\usepackage{graphicx}
\graphicspath{ {./} }

\begin{document}
\maketitle
\begin{abstract}

News articles such as sports game reports are often thought to closely follow the underlying game statistics, but in practice they contain a notable amount of background knowledge, interpretation, insight into the game, and quotes that are not present in the official statistics. This poses a challenge for automated data-to-text news generation with real-world news corpora as training data. We report on the development of a corpus of Finnish ice hockey news, edited to be suitable for training of end-to-end news generation methods, as well as demonstrate generation of text, which was judged by journalists to be relatively close to a viable product. The new dataset and system source code are available for research purposes.\footnote{\url{https://github.com/scoopmatic/finnish-hockey-news-generation-paper}}

\end{abstract}

\section{Introduction}

Automated, or robotic, journalism aims at news generation from structured data sources, either as the final product or as a draft for subsequent post-editing. At present, automated journalism typically focuses on domains such as sports, finance and similar statistics-based reporting, where there is a commercial product potential due to the high volume of news, combined with the expectation of a relatively straightforward task.

News generation systems---especially those deployed in practice---tend to be based on intricate template filling, aiming to give the users the full control of the generated facts, while maintaining a reasonable variability of the resulting text. This comes at the price of having to develop the templates and specify their control logic, neither of which are tasks naturally fitting journalists' work. Further, this development needs to be repeated for every domain, as the templates are not easily transferred across domains. Examples of the template-based news generation systems for Finnish are Voitto\footnote{\url{https://github.com/Yleisradio/avoin-voitto}} by the Finnish Public Service Broadcasting Company (YLE) used for sports news generation, as well as Vaalibotti~\cite{leppanen-etal-2017-data}, a hybrid machine learning and template-based system used for election news.

\citet{wiseman-etal-2018-learning} suggested a neural template generation, which jointly models latent templates and text generation. Such a system increases interpretability and controllability of the generation, however, recent sequence-to-sequence systems represent the state-of-the-art in data-to-text generation.~\cite{dusek-2018-e2e}

In this paper, we report on the development of a news generation system for the Finnish ice hockey news domain, based on sequence-to-sequence methods. In order to train such a system, we compile a corpus of news based on over 2000 game reports from the Finnish News Agency STT. While developing this corpus into a form suitable for training of end-to-end systems naturally requires manual effort, we argue that compiling and refining a set of text examples is a more natural way for journalists to interact with the system, in order for them to codify their knowledge and to adapt it for new domains.

Our aim is to generate reports that give an overview of a game based on information inferrable from the statistics. Such reports can be used either as a basis for further post-editing by a journalist imprinting own insights and background information, or even used directly as a news stream labelled as machine-generated.

In the following, we will introduce the news dataset and the process of its creation, introduce an end-to-end model for news generation, and evaluate its output respective to the abovementioned objectives.

\section{Ice Hockey News Dataset}

An ice hockey game is recorded into statistics in terms of different events occurring during play, such as goals and penalties. In order to train a model to generate game reports, we need access to these events, as well as example news articles about the game. Only recently have game statistics become available to the public through a web interface or API, whereas the information has traditionally been recorded as structured text files.

The news corpus from the Finnish News Agency STT\footnote{A version of the corpus is available at \url{http://urn.fi/urn:nbn:fi:lb-2019041501} for academic use.} includes, among all other news, articles covering ice hockey games in the Finnish leagues during the years 1994--2018. In addition to news articles, the corpus also includes the original game statistics text files. This creates an opportunity to align the game statistics with the corresponding news articles, producing a dataset of over 20 years of ice hockey data with reference news articles for the games. When automatically pairing the game statistics with news articles using date and team names as a heuristic, we obtain a total of 3,454 games with existing statistics and at least one corresponding news article.

Utilizing real journalistic material poses a challenge in that the articles mix information that can be found directly in the game statistics (e.g., scores and names) with  information inferable from the statistics (e.g., statements such as \emph{shortly after}), information based on background knowledge (e.g., a team's home city or player's position), game insight and judgement based on viewing the game (e.g., expressions such as \emph{slapshot} or \emph{tipping the puck} describing the character of a shot), and even player interviews. 

Therefore, directly using the limited amount of actual news articles for end-to-end system training becomes problematic. In our initial experiments the generation model learns to ``hallucinate'' facts, as easily occurs when the target text is too loosely related to the conditioning input.\footnote{This observation is also supported by \citet{wiseman-etal-2017-challenges} mentioning that their generation model occasionally ``hallucinates factual statements'' that are plausible but false.} In order to ensure that the generation model is able to learn to generate accurate descriptions from game statistics, we clean the news corpus by manually aligning corresponding text spans with game events detailed in the statistics.

For the sake of comparison, let us consider the Rotowire corpus \cite{wiseman-etal-2017-challenges} containing basketball game summaries and statistics, which was recently released and has become a popular data set for training data-to-text generation systems (cf., e.g., \citet{nie-etal-2018-operation,wiseman-etal-2018-learning,puduppully2019data}). 
The Rotowire game summaries are straightforward in their style of reporting, focusing on the game at hand and tend for the most part to reference facts in the statistics. By contrast, our news corpus is more heterogeneous, including both articles focusing on the particular game and articles that take a broader perspective (e.g., describing a player's career). The STT news articles tend to read in the journalist's voice, putting substantial emphasis on the character of the game, often in colorful language, as well as quoting players and coaches.

\begin{figure*}
\scriptsize
\centering
\includegraphics[width=\textwidth]{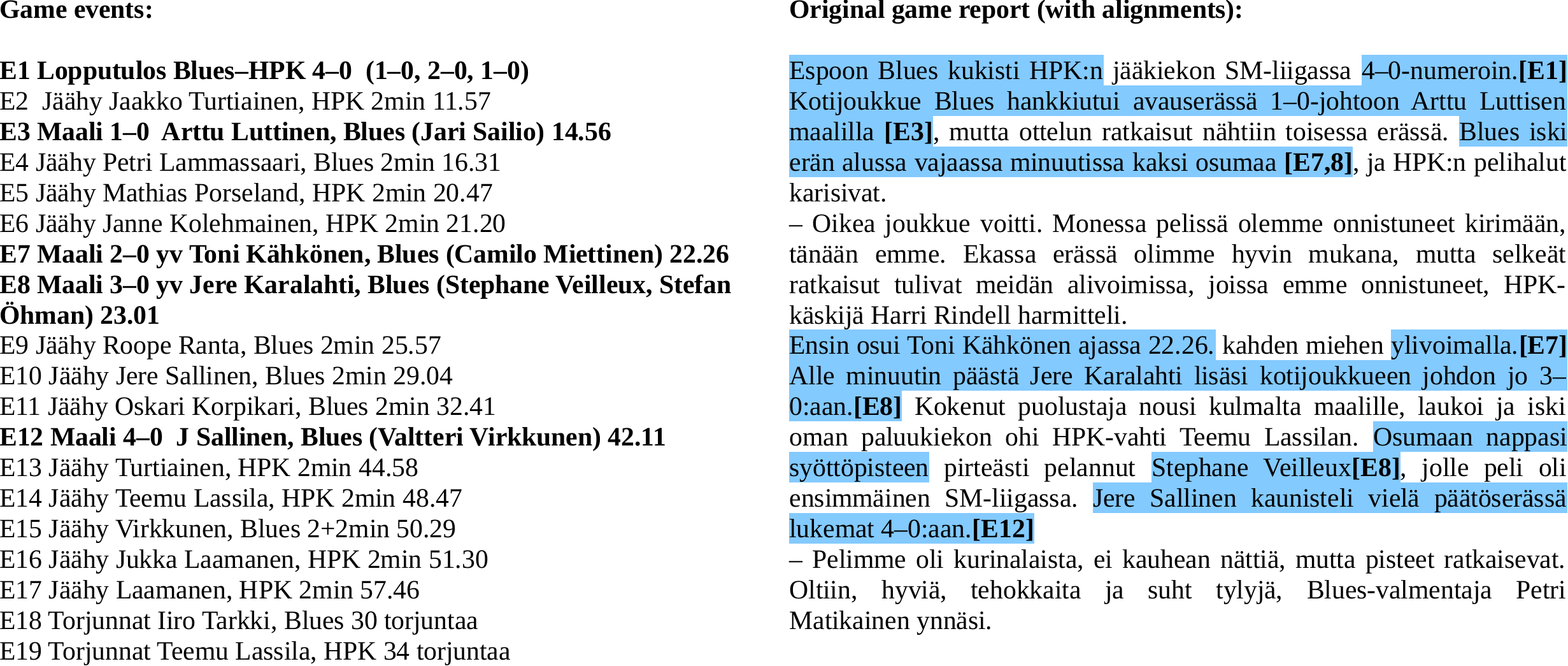}
\caption{A representative example of events extracted from game statistics and the corresponding report in the news corpus. Events that are mentioned in the article are boldfaced (left) and the aligned text spans are highlighted in blue (right). Event references are indicated at the end of each alignment (e.g., [E1]).}
\label{fig:alignment}
\end{figure*}

An example of the events available in the game statistics, the actual news article on the game, and how these align, is shown in Figure~\ref{fig:alignment}. Text spans highlighted with blue color are based on information available in the statistics, all other being external information. It illustrates the typical portion of a raw article that is not inferrable from the data. English translations are available for a comparable example in Figure~\ref{fig:generation-example}.

\subsection{Extraction of Game Events}
\label{sec:extraction}

For each event occurring in a game and recorded in statistics, we identify its type and associated features. There are four event types: \emph{end result}, \emph{goal}, \emph{penalty} and \emph{save}. As a general rule, for each game the end result entry specifies total scores, participating teams and additional circumstances of the game such as overtime or shootout. The goal event is the most frequent and includes features such as goal scorer, assists, team, resulting score, time and the current state of the game (e.g., power play, penalty shot). We also derive special features that in many cases require consideration of other events in the context, but pertain to one particular event, e.g., is the goal deciding or final. The penalty event specifies player, team, time in the game and penalty time. The save event summarises the number of saves of a goaltender/team. 

We perform information extraction with regular expressions on the statistics in order to structure the game into a chronological sequence of events, which can then be aligned with the corresponding text spans from the news article, as well as used as input in the text generation. An initial sequence of events after information extraction is shown on the left side in Figure~\ref{fig:alignment}. Before text generation, these events are yet enhanced with additional information derived from the event itself, or the game.


\subsection{Manual Alignment}

The manual alignment of game events and the article text is carried out by a single annotator in approximately 6 weeks as an integral part of the system development work. The annotator receives a sequence of extracted events and aligns these with the corresponding news article, typically but not necessarily a single sentence for every expressed event. All ungrounded parts within the text are removed and if necessary, the text is modified for fluency. We cannot align full sentences with events as, for instance, \citet{barzilay-lapata-2005-collective} do, as often information not grounded in the statistics is expressed together with statistics-based facts within the same sentences and clauses.


The alignment process therefore frequently requires rephrasing the text, for instance, in order to make it more neutral and avoid generating arbitrary subjective judgements. We find that the news article commonly includes description that is not evident from the data (e.g., subjective characteristics of the player or the shot), and often may reflect the reporter's viewpoint. For instance, the reporter may evaluate an event negatively by writing \emph{Player A did not manage to score more than one goal}, reflecting expectation of the player's performance, which we would change into a more objective form such as \emph{Player A scored one goal}. Similarly, word choices based on viewing the game (e.g., \emph{slapshot}) are changed to more neutral counterparts.

After removing the uninferable parts of a sentence, the remainder is often relatively short, in which case we sometimes opt to replace the information that was removed with references to other appropriate information from the game statistics, such as the time of the event or the state of the game. This serves to maintain a similar sentence and clause structure as that of the original corpus.

In the majority of our aligned data one text span refers to one game event. However, in some cases the same indivisible span can describe multiple events, most commonly of the same type (e.g., \emph{Player A scored two goals}, or \emph{Team A received a total of eight penalties}). In such cases, we produce multi-event alignments, where all related events are aligned with a single text span.\footnote{ Cf. Figure~\ref{fig:alignment}, alignments of E7 and E8: The events are expressed differently depending on type of alignment, where the 2-to-1 aligned text says that the team scored two goals.} Multi-event alignments support the development of more advanced generation, which stands to produce more natural-sounding reports in terms of less repetition and more flexible structuring.

\subsection{Corpus Statistics}
\label{sec:corpus_stats}


In Table~\ref{tbl:hockey-corpus}, we summarize the overall size statistics of the final ice hockey corpus after the statistics and news articles have been automatically paired, and events have been manually aligned with the text. In total, 2,307 games were manually checked (66.8\% of the paired corpus), of which 2,134 games were correctly paired with the article describing the game. In these games, 12,251 out of 36,097 events (33.9\%) were referenced in the text and successfully aligned. \emph{End result} occurs in nearly all news articles as the first event, whereas the \emph{goal} event is by far the most frequent one, each game mentioning on average 3.1 goals. The number of aligned events is greater than the number of aligned text spans due to multi-event alignments. While 82.1\% of text spans align with a single event, there is a long tail of multi-event alignments, with 11.2\% aligning to two events, 3.4\% to three, 1.4\% to four, etc.

\begin{table}[t]
\centering
\begin{tabular}{l|r}
 Games & 2,134 \\ 
 Events & 36,097 \\ 
 Aligned events & 12,251 \\
 \hspace{3mm} End result & 2,092 \\ 
 \hspace{3mm} Goal & 6,651 \\ 
 \hspace{3mm} Penalty & 2,163 \\ 
 \hspace{3mm} Save & 1,345 \\ 
 Aligned spans & 8,831 \\
 Aligned sentences & 9,266 \\
 Aligned tokens & 84,997\\
 
\end{tabular}
\caption{Event alignment statistics.}
\label{tbl:hockey-corpus}
\end{table}

%

\begin{table}[t]
\centering
\begin{tabular}{l|cc}
                  & Original Corpus & Aligned \\\hline
Sentences         & 49,496               & 9,266               \\
Tokens            & 601,990              & 84,997                \\
Unique tokens     & 54,624               & 7,393                \\
Unique lemmas     & 24,105               & 4,724                \\
STTR (words) & 0.641                & 0.525                 \\
STTR (lemmas)& 0.501                & 0.424                 
\end{tabular}
\caption{Comparison of the original hockey news and the aligned sentences, including analysis of lexical diversity.}
\label{tbl:lexical}
\end{table}

In Table~\ref{tbl:lexical} we measure the lexical diversity of original ice hockey news articles, as well as the resulting dataset after manual alignment, by computing the Standardized Type--Token Ratio (STTR). The measure is defined as the number of unique tokens or lemmas divided by the total number of tokens, calculated on every segment of 1,000 tokens separately and averaged across the corpus. STTR is more meaningful than the standard type--token ratio when comparing corpora of substantially different sizes. Both corpora are tokenized and lemmatized using the Turku Neural Parser pipeline~\cite{udst:turkunlp,kanerva2019universal}. STTR of the aligned corpus is lower than in the original hockey news on both word and lemma level, indicating a somewhat---but not substantially so---more restricted vocabulary use in our aligned subset.


\section{Event Selection Model}
\label{sec:selection-model}

As illustrated previously, any given news article describes only a subset of most noteworthy events of a game. We observe that most reports are concise, referencing on average 5.7 events. 
The distribution of events/report is: 1\textsuperscript{st} quantile at 3 events (20.9\% of events in game), 2\textsuperscript{nd} at 5 (22.2\%), 3\textsuperscript{rd} at 7 (38.5\%), 4\textsuperscript{th} at 36 (100\%).

Our alignment serves as a gold standard reflecting which events the journalists have chosen to mention for each game. In our generation task, we are presented with the problem of selecting appropriate events from the full game statistics. We use the gold standard selection during training and validation of the text generation model, as well as the automatic evaluation. As we deploy our text generation model for manual evaluation, we use a Conditional Random Field (CRF) model to predict which events to mention.

Casting the event selection problem as a sequence labeling task, the CRF model takes as input the full sequence of events in one game together with associated features for each event, and predicts a binary label for each event.\footnote{We use label weighting to account for the imbalanced distribution, which we optimize against the validation set to 0.85:1 for the positive class (other optimal hyperparameters are C1=35.0, C2=0.5, as well as defaults). We use CRFsuite \cite{CRFsuite} with label weighting by Sampo Pyysalo: \url{https://github.com/spyysalo/crfsuite}} We achieve an overall F-score of 67.1\% on the test set, which broken down by event type is: end result (98.0\%), goal (70.2\%), penalty (20.1\%), save (47.7\%). Penalties are the most difficult to predict, being reported only 7.8\% of the time in reality, e.g., compared to 54.1\% for goals. 

\section{Text Generation}

Next, we present the model architecture used in text generation, and evaluate the model on a popular baseline dataset. After that, we describe the training of the generation model on our ice hockey corpus and use automatic evaluation metrics to compare against existing references.

\subsection{Model Architecture}

We use a pointer-generation network~\cite{vinyals2015pointer,gu2016incorporating,see-2017-get}, where the neural attention mechanism in the encoder-decoder model is adapted to jointly model a probability distribution over words from the known vocabulary, a distribution over words from the input sequence to copy and a probability that controls the copying mechanism. A separate coverage attention vector, a sum of past attention distributions, is maintained to inform the model of its past attention decisions. Such a coverage model is shown to prevent text repetition in generated output~\cite{tu2016modeling,see-2017-get}.

The model is implemented using the OpenNMT-py library~\cite{opennmt}. The encoder has two bidirectional LSTM layers with 500 hidden units, together with 500-dimensional word embeddings. The decoder has two unidirectional LSTM layers with 500 hidden units. Both encoder and decoder apply a dropout of 0.3 between LSTM layers.

\subsection{Baseline Experiments on the E2E Dataset}

\begin{table*}[ht!]
\centering
\begin{tabular}{c|lllll}
       & BLEU    & NIST    & METEOR  & ROUGE-L & CIDEr  \\\hline
Ours   & 0.6758  & 8.5588  & 0.4536  & 0.6990  & 2.2007 \\
TGen   & 0.6593  & 8.6094  & 0.4483  & 0.6850  & 2.2338 \\
ST top & 0.6619* & 8.6130* & 0.4529** & 0.7083$\dag$ & 2.2721$\ddag$ \\\hline
Ours single ref. & 0.3190  & 5.1995   & 0.3574   & 0.4969   & 1.7922   \\
\end{tabular}
\caption{Performance of our generation model on the E2E test set compared to the shared task baseline (TGen) and winners on each metric (*\citet{juraska-2018-deep}, **\citet{puzikov-2018-e2e}, $\dag$\citet{zhang-2018-e2e}, $\ddag$\citet{gong-2018-e2e}), as well as our model in an adapted evaluation setup (Ours single ref.).}
\label{tbl:restaurant_results}
\end{table*}

To demonstrate the performance of our generation model architecture, we report results on a known dataset with published baselines, namely the E2E NLG Challenge \cite{dusek-2018-e2e} on end-to-end natural language generation in spoken dialogue systems. The task is to produce a natural language description of a restaurant based on a given meaning representation (MR)---an unordered set of attributes and their values. The attributes included, among others, the restaurant name, area, food type and rating. We represent the given MR as a sequence of tokens where each attribute value is embedded into XML-style beginning and end attribute markers, and the order of attributes is kept fixed across the whole dataset. The target output is a sequence of tokens. We do not apply any explicit delexicalization steps.

In Table~\ref{tbl:restaurant_results} we measure BLEU \cite{Papineni-2002-bleu}, NIST \cite{Doddington-2002-nist}, METEOR \cite{lavie-2007-meteor}, ROUGE-L \cite{lin-2004-rouge} and CIDEr \cite{Vedantam-2015-CIDEr} metrics on the 2018 E2E NLG Challenge test data using the evaluation script provided by the organizers\footnote{\url{https://github.com/tuetschek/e2e-metrics}}. Our generation system is compared to the official shared task baseline system, TGen \cite{dusek-jurcicek-2016-sequence}, as well as to the top performing participant system on each score (ST top). Our system outperforms the TGen baseline on 3 out of 5 metrics (BLEU, METEOR and ROUGE-L), which is on par with the official shared task results, where not a single one participant system was able to surpass the baseline on all five metrics. On two metrics, BLEU and METEOR, our system outperforms the best shared task participants.

E2E NLG Challenge evaluation is based on having multiple references for each MR, on average each unique MR in the corpus having 8 reference descriptions. In the evaluation, the output for each unique MR is compared against all references and the maximum score is used, naturally leading to higher scores. To have more comparable numbers to our ice hockey corpus, where we have only one reference for each input event, we also include scores obtained by comparing each MR to each of its reference descriptions separately as if they were individual data points (Ours single ref.).



\subsection{Hockey Data Representation}

Currently, we concentrate on training the model only with text spans aligning with single events, excluding the less frequent multi-event alignments. Furthermore, we are considering each event as a separate training example, independent of other events in the game.

Given a single event described as a sequence of features and their values, our text generation model is trained to produce the text span aligned with it. Following the data representation used in E2E NLG Challenge experiments, the input events are represented as a linearized sequence of tokens, where XML-style beginning and end tags are used to separate the different features (see Figure~\ref{fig:train-data}). This allows the model to directly copy some of the input tokens to the output when necessary. The ability to copy tokens is especially important with player names and exact times, where the vocabulary is sparse, and many of these can even be previously unseen, unknown tokens. In addition, we also include features that are meant to inform generation without being copied themselves, for example, the type of the event. The target of generation is a tokenized sequence of words, where also dashes inside scores are separated, allowing the model to swap scores when necessary. The reference text sometimes flips 
the order of the teams, requiring the score to be inverted as well (\emph{Team A--Team B 0--1} into \emph{Team B won Team A 1--0}).

\begin{figure*}[ht!]
\scriptsize
\centering
\begin{BVerbatim}

INPUT:   <length>long</length> <type>result</type> <home> Ässät </home> <guest> Blues </guest>
         <score> 0 - 4 </score> <periods> ( 0 - 3 , 0 - 0 , 0 - 1 ) </periods>
OUTPUT:  Blues vei voiton Ässistä maalein 4 - 0 ( 3 - 0 , 0 - 0 , 1 - 0 ) .
\end{BVerbatim}
\caption{An example input--output pair for the text generation model, derived from manual alignment. The original, untokenized output sentence is \emph{Blues vei voiton Ässistä maalein 4--0 (3--0, 0--0, 1--0).} \\
(Literal English translation: \emph{Blues took a win from Ässät with goals 4--0 (3--0, 0--0, 1--0).})}
\label{fig:train-data}
\end{figure*}

One particular challenge in our corpus is that the decision of which aspects to focus on, i.e., which particular features from the source event to verbalize, is relatively arbitrary. For example, sometimes a journalist mentions the player or players assisting a goal, but in many cases it is left out. Both options are equally correct, but overlap-based metrics such as BLEU penalize such creative variation. By contrast, in other text generation datasets such as the E2E NLG Challenge, 
the output text in general describes all input features. To account for this variation, we include a length feature to act as a minimal supervision signal for the model to rely on. We divide output text lengths into three evenly sized categories (\emph{short}, \emph{medium} and \emph{long}) to provide a hint to the model during training of how long and detailed output it is expected to generate for each training example. At test time, we then have the possibility to control the desired approximate length of the generated text. In the experiments throughout this paper, we generate all three length variants for each event and pick the one with the highest average confidence of the generation model. 

\subsection{Training and Optimization}

The model is trained using the Adam optimizer with learning rate of 0.0005 and batch size of 32. The model is trained for a maximum of 8000 steps (ca.\ 40 epochs), and the final model is chosen based on validation set performance. We use 80\% of the aligned games for training, 10\% for validation and 10\% for testing.

In our initial experiments, we used the RBFOpt library~\cite{costa2014rbfopt} for hyperparameter tuning, maximizing validation set BLEU score. However, when manually inspecting generation results on the validation set, we noticed that models with a higher BLEU score result in more fluent text but generate more factual mistakes. This observation is supported by \citet{wiseman-etal-2017-challenges}, who note that BLEU score tends to reward fluent text rather than other aspects desirable in generation of sports summaries from data. For this reason, we ultimately decided to use hyperparameters manually tuned on the validation set to give a good perceived balance between fluent and factually correct text.

\subsection{Automatic Evaluation of Hockey Generation}

In Table~\ref{tbl:hockey-bleu}, we provide evaluation results using the five aforementioned metrics. We evaluate on event level using gold standard event selection, where each generated event description is compared to its existing reference text. As the model is trained to produce a tokenized sequence, we apply a detokenizer to be able to compare against the original untokenized reference text. On the test set, the model achieves a BLEU score of 19.67. To the extent that different datasets allow comparison, the best reported score on the Rotowire basketball news corpus is 16.50 \cite{puduppully2019data}. Compared to our earlier E2E baseline experiment, we score lower than our closest comparable reference of 31.90 (with single references), which is understandable due to the much smaller train set size for the hockey corpus (about 13\% in size).

\begin{figure}
    \centering
    \includegraphics[width=\columnwidth,height=5.05cm]{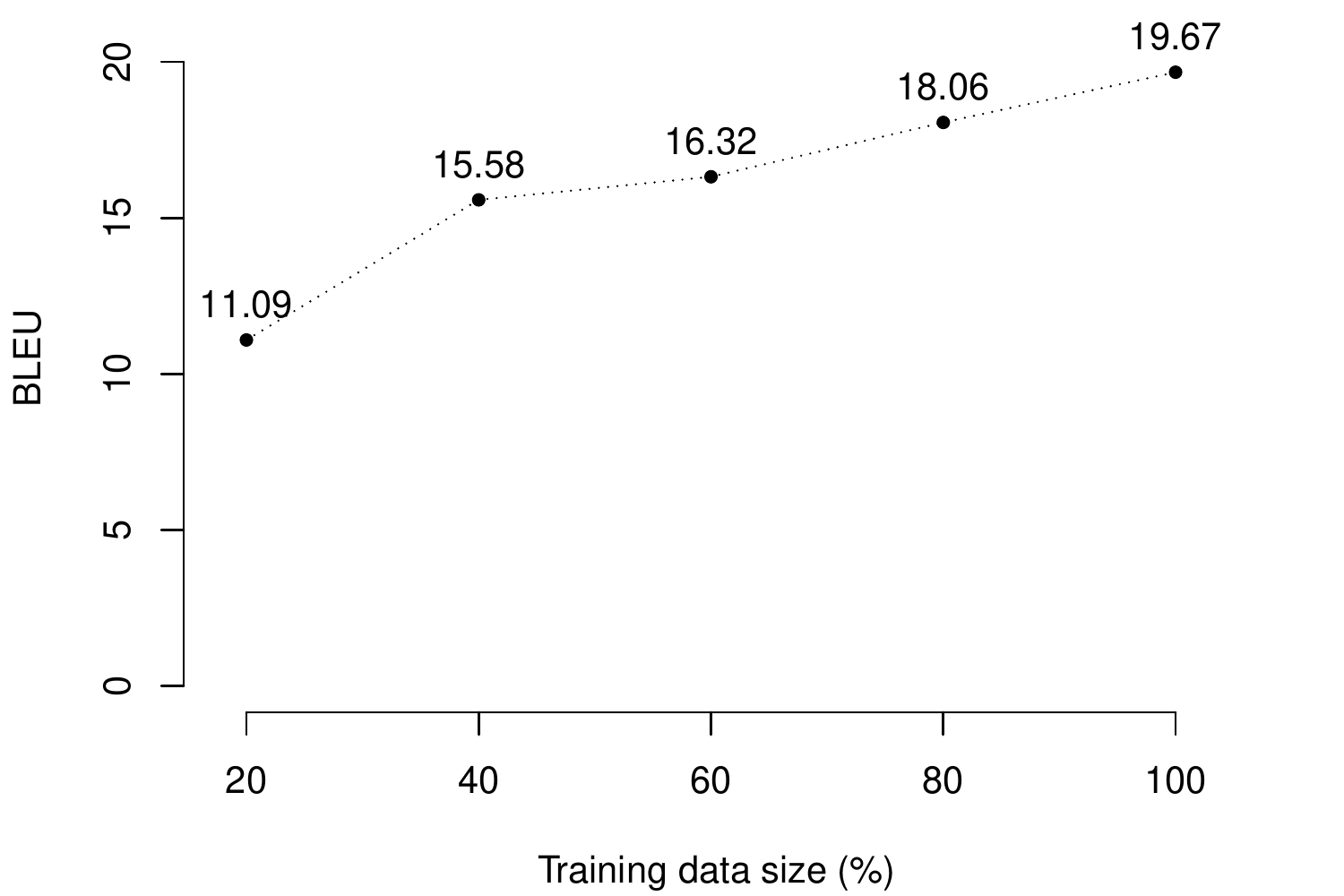}
    \caption{Learning curve demonstrating the effect of training data size on generation performance.}
    \label{fig:learning-curve}
\end{figure}

In Figure~\ref{fig:learning-curve}, we plot the learning curve with increasing sizes of training data in order to illustrate how generation performance benefits from more data. The learning curve is still steadily increasing when using 100\% of the training data currently available, which indicates that more data would most likely further improve the performance.


\begin{table}
    \centering
    \begin{tabular}{l|c}
        BLEU & 0.1967 \\
        NIST & 4.4144 \\
        METEOR & 0.2297 \\
        ROUGE-L & 0.4159 \\
        CIDEr & 1.8658 \\
    \end{tabular}
    \caption{Automatic evaluation metrics on hockey corpus test set.}
    \label{tbl:hockey-bleu}
\end{table}

\section{Human evaluation}

As our objective is practically usable news generation, we carry out a manual evaluation of its output on 59 randomly selected games from the test set, focusing in particular on the corrections that would be necessary to obtain acceptable output. Example corrections are shown in Figure~\ref{fig:generation-example}.
These full game reports are generated by first applying the selection model described in Section~\ref{sec:selection-model} to select the events to be included in the report, and then using the text generation model to verbalize each selected event. The generated texts of the events are then detokenized and concatenated in chronological order into a full game report.

\begin{table}[ht]
    \centering
    \begin{tabular}{l|r}
        Error type & Count\\\hline
        Team or player name & 25 \\
        Type or score of goal & 24\\
        Time reference & 14 \\
        Total score & 6 \\
        Penalty & 5 \\
        Assist & 2 \\
        Power play & 2
    \end{tabular}
    \caption{Types of factual errors in the generated output for 59 games.}
    \label{tab:errortypes}
\end{table}

\subsection{Minimum Edit Evaluation}

In the \emph{minimum edit} evaluation, carried out by the annotator who created the news corpus, only factual mistakes and grammatical errors are corrected, resulting in text which may remain awkward or unfluent. The word error rate (WER) of the generated text compared to its corrected variant as a reference is 5.6\% (6.2\% disregarding punctuation). The WER measure is defined as the number of insertions, substitutions, and deletions divided by the total length of the reference, in terms of tokens. The measure is the edit distance of the generated text and its corrected variant, directly reflecting the amount of effort needed to correct the generated output.

\subsubsection{Factual Correctness}

The factual errors and their types are summarized in Table~\ref{tab:errortypes}. From the total of 510 game events generated by the system, 78 of these contained a factual error, i.e.\ 84.7\% were generated without factual errors.

The most notable errors involved player or team names. The input for a single game event may contain more than one name (e.g.\ the goal scorer and up to two assisting players). In these cases, the model occasionally paired first name and surname incorrectly. Less frequent errors include wrong team pairings when referring to the end result of the game, e.g., \emph{Team A lost to Team A}.

In sentences where the exact time of a game event is generated as a number (\emph{at time 39.54}) the model copied the time reliably, but when the approximate time of an event is verbalized (\emph{at the end of the second period}) there were occasional errors. Another notable error category is types of goals or their scores (e.g.\ 3--0, deciding goal, tying goal). In this category, the error similarly occurred in cases when the reference is verbal (\emph{third goal}), but occasionally also in numerical representation (\emph{3--0} instead of \emph{0--3}). The other, less common categories relate to game total scores, power play, assists and penalties.

\subsubsection{Fluency}

Overall, the generated text is highly grammatical. The most frequent grammatical error in the output is unconjugated player names; commonly names in their Finnish nominative case that should be in genitive. In a few cases, the model makes grammatical errors when it copies an incompatible word from the input, e.g., a name instead of a time reference.

Both error types commonly occur when the model copies from the input. As it operates on the token rather than sub-word level, it is challenging to map between word forms and inflect infrequent words such as player names. The latter error likely occurs when the copy attention is indecisive and fails to recognize the type for infrequent tokens.

Most fluency issues relate to the overall flow and structure of the report. Addressing these issues would require the model to take into account multiple events in a game, and combine the information more flexibly to avoid repetition. For instance, the output may repeatedly mention the period number for all goals in the same period. Likewise, this setup sometimes results in unnatural, yet grammatical, repetition of words across consecutive sentences. Even though the model has learned a selection of verbs meaning \emph{to score a goal}, it is unable to ensure their varied use. While not successful in our initial experiments, generating text based on the multi-event alignments or at document level may eventually overcome these issues.





\begin{figure*}[ht!]
    \centering
    \includegraphics[width=\textwidth]{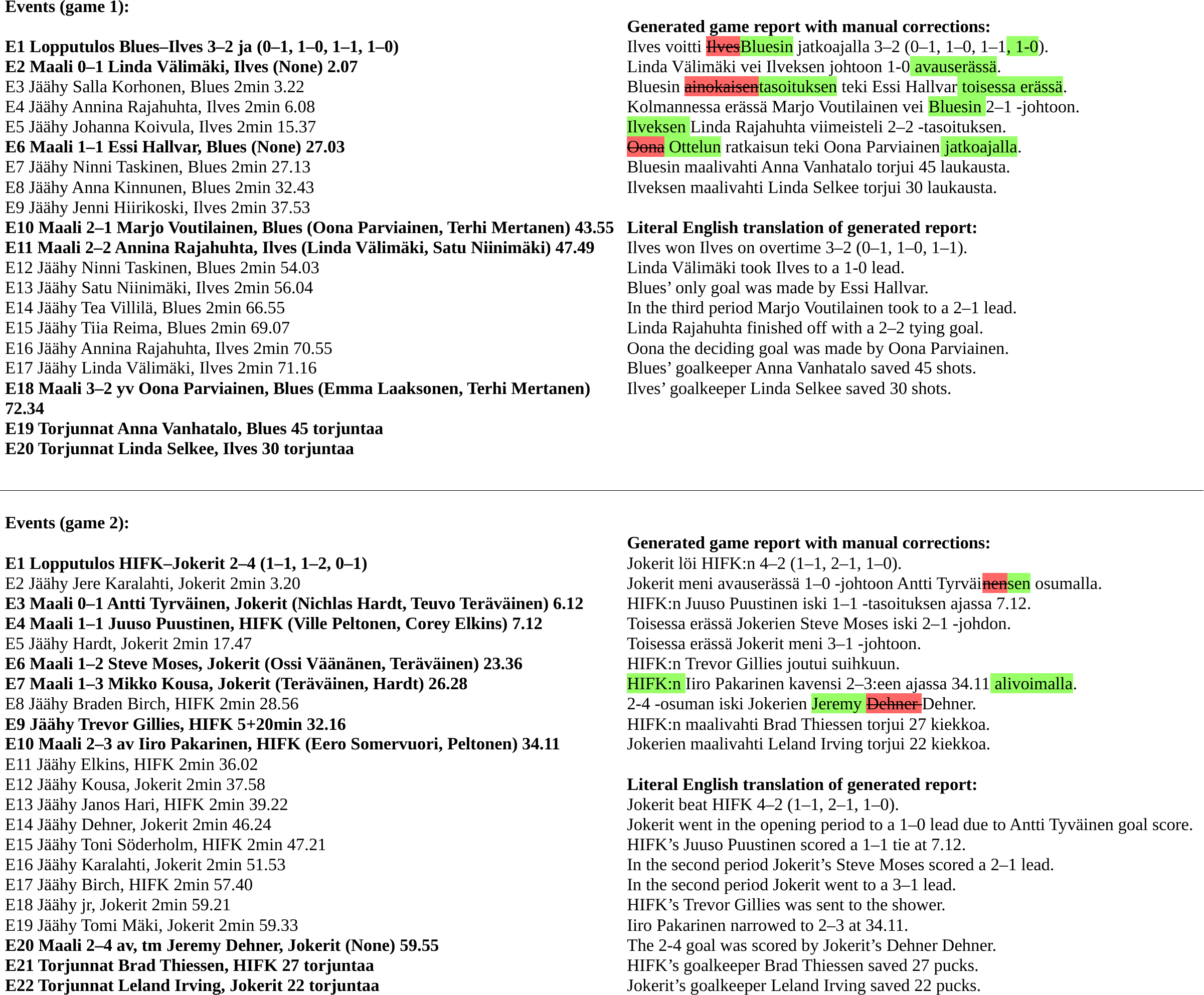}
    \caption{Generated reports with manual corrections for two example games. Insertions in manual corrections are marked in green and deletions in red and struck through. English translations are based on original, uncorrected generation output. English translations for input events: Lopputulos (End result), Maali (Goal), Jäähy (Penalty), Torjunnat (Saves), ja (overtime), yv (power play), av (short-handed), tm (empty net).}
    \label{fig:generation-example}
\end{figure*}

\subsection{Product-Readiness Evaluation}
\label{sec:product-readiness}

The second human evaluation aimed at judging the acceptability of the output for production use in a news agency. The output is evaluated in terms of its usability for a news channel labelled as being machine-generated, i.e.\ not aiming at the level of a human journalist equipped with substantial background information. The evaluation was carried out by two journalists from the STT agency, who split the 59 games among themselves approximately evenly. The first journalist edited the games to a form corresponding to a draft for subsequent minor post-editing by a human, simulating the use of the generated output as a product where the final customer is expected to do own post-editing before publication. The second journalist directly edited the news to a state ready for direct publication in a news stream labeled as machine-generated news. In addition to correcting factual errors, the journalists removed excessive repetition, improved text fluency, as well as occasionally included important facts which the system left ungenerated. The WER measured against the output considered ready for post-editing, is 9.9\% (11.2\% disregarding punctuation), only slightly worse than the evaluation with only the factual and grammatical errors corrected. The WER measured against the output considered ready for direct release, was 22.0\% (24.4\% disregarding punctuation). In other words, 75--90\% of the generated text can be directly used, depending on the expected post-editing effort.

Figure \ref{fig:generation-example} shows two example games along with the generated reports and manual corrections made by the journalist in order to prepare it for publication. Literal translations from the generated, uncorrected Finnish into English are provided for reference.




\section{Conclusions and Future Work}

We developed and evaluated an end-to-end system for news generation from structured data, using a corpus of news and game statistics in the ice hockey domain. In terms of the data, our primary finding was the level to which professionally produced news contain information that cannot be inferred from the game statistics. This leads to the model learning to "hallucinate" facts and necessitates a manual alignment and editing of the training data. Once we created a suitable training dataset, we were able to generate highly grammatical text which, in terms of word error rate (edit distance), was relatively close to what was judged as a viable product by domain journalists. We found that most factual errors in the generated output fall into a small number of categories, mostly related to copying names from the input, types of events, and time references. Addressing these errors is a matter of future work and can be approached using data augmentation techniques as well as introducing sub-word units which would allow the model to deal with inflections.

Currently, we only generate the news as independent events, with roughly one sentence corresponding to one event. As this leads to a somewhat unnatural text, we have attempted in preliminary experiments to generate whole news texts at once, as well as sentences combining several events, nevertheless with results far from useful. This is likely due to the relatively small number of training examples where a single sentence accounts for several distinct events. We will focus on this problem in our future work, investigating methods which would allow pre-training the generation model so as to be able to successfully accept several events on its input.

The new dataset, the original news corpus and the source code of the model are available for research use. \footnote{\url{https://github.com/scoopmatic/finnish-hockey-news-generation-paper}}



\section*{Acknowledgments}

We gratefully acknowledge the collaboration of Maija Paikkala, Salla Salmela and Pihla Lehmusjoki of the Finnish News Agency STT, as well as the support of the Google Digital News Innovation Fund, Academy of Finland, CSC -- IT Center for Science, and the NVIDIA Corporation GPU Grant Program.


\bibliographystyle{acl_natbib}
\bibliography{nodalida2019}

\end{document}